\documentclass[10pt,twocolumn,letterpaper]{article}
\PassOptionsToPackage{table,xcdraw}{xcolor}
\usepackage[accsupp]{axessibility} 
\usepackage{algorithm}     
\usepackage{algpseudocode} 
\usepackage{cvpr}              
\usepackage{graphicx}
\usepackage{caption}  
\usepackage{amsmath}
\usepackage{amssymb}
\usepackage{booktabs}
\usepackage{enumitem}
\usepackage[table,xcdraw]{xcolor} 
\usepackage{graphicx}             
\usepackage{booktabs} 

\usepackage{multirow}

\definecolor{cvprblue}{rgb}{0.21,0.49,0.74}
\usepackage[pagebackref,breaklinks,colorlinks,allcolors=cvprblue]{hyperref}
\usepackage{float}
\def\paperID{30321} 
\def\confName{CVPR}
\def\confYear{2026}

\title{Denoise and Align: Towards Source-Free UDA for Robust Panoramic Semantic Segmentation}

\author{
Yaowen Chang$^{1,*}$ \quad 
Zhen Cao$^{1,}$\thanks{Equal Contribution}\quad 
Xu Zheng$^{2}$ \quad 
Xiaoxin Mi$^{3}$ \quad 
Zhen Dong$^{1,}$\thanks{Corresponding author}\\
$^{1}$Wuhan University \quad $^{2}$HKUST (Guangzhou) \quad $^{3}$Wuhan University of Technology \\
{\tt\small \{yaowenchang, zhen.cao, dongzhenwhu\}@whu.edu.cn, zhengxu128@gmail.com, xiaoxin.mi@whut.edu.cn}
}

\begin{document}
\twocolumn[{%
\renewcommand\twocolumn[1][]{#1}%
\maketitle
\begin{center}
    \vspace{-20px}
    \includegraphics[width=0.9\textwidth, trim=0cm 0cm 0cm 0.4cm, clip]{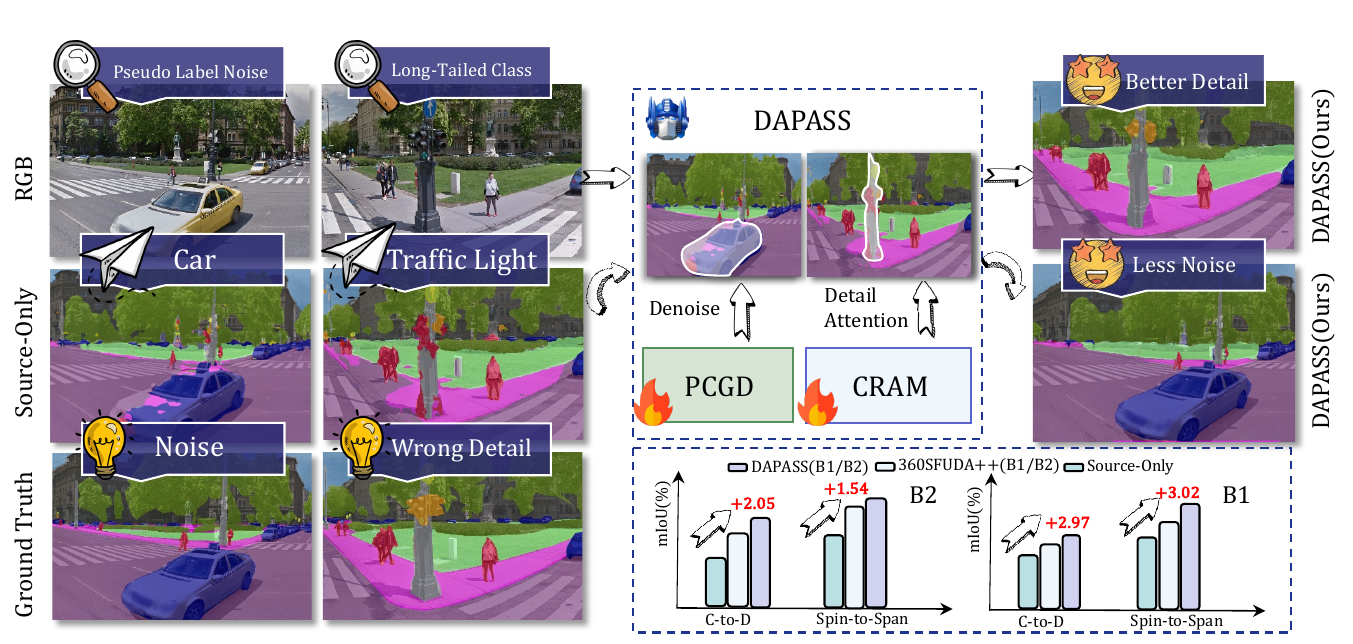}
    \vspace{-5px}
    \captionof{figure}{Teaser of our DAPASS. Compared with source-only (Only use the pinhole source domain image datasets train the model without adaptation) and existing SOTA SFUDA methods~\cite{zheng2024360sfuda++}, DAPASS effectively removes pseudo-label noise and recovers fine details for different classes, benefiting from its PCGD and CRAM modules. Performance improvements are shown across cross-domain tasks.}
    \vspace{-5px}
    \label{fig:Teaser}
\end{center}
}]

{
  \renewcommand{\thefootnote}{*}
  \footnotetext[1]{Equal Contribution}
  \footnotetext[2]{Corresponding author}
}

\begin{abstract}
\vskip -2.0ex
Panoramic semantic segmentation is pivotal for comprehensive 360° scene understanding in critical applications like autonomous driving and virtual reality. However, progress in this domain is constrained by two key challenges: the severe geometric distortions inherent in panoramic projections and the prohibitive cost of dense annotation. While Unsupervised Domain Adaptation (UDA) from label-rich pinhole-camera datasets offers a viable alternative, many real-world tasks impose a stricter source-free (SFUDA) constraint where source data is inaccessible for privacy or proprietary reasons. This constraint significantly amplifies the core problems of domain shift, leading to unreliable pseudo-labels and dramatic performance degradation, particularly for minority classes. To overcome these limitations, we propose the DAPASS framework. DAPASS introduces two synergistic modules to robustly transfer knowledge without source data. First, our Panoramic Confidence-Guided Denoising (PCGD) module generates high-fidelity, class-balanced pseudo-labels by enforcing perturbation consistency and incorporating neighborhood-level confidence to filter noise. Second, a Cross-Resolution Attention Module (CRAM) explicitly addresses scale variance and distortion by adversarially aligning fine-grained details from high-resolution crops with global semantics from low-resolution contexts. DAPASS achieves state-of-the-art performances on outdoor (Cityscapes-to-DensePASS) and indoor (Stanford2D3D) benchmarks, yielding 55.04\% (+2.05\%) and 70.38\% (+1.54\%) mIoU, respectively. Code and models are publicly available at \href{https://github.com/ZZZPhaethon/DAPASS}{https://github.com/ZZZPhaethon/DAPASS}. 
\end{abstract}
\vskip-3ex
    
\vskip-3ex
\section{Introduction}
\label{sec:intro}
As autonomous driving perception systems \cite{ma2021densepass,yang2021capturing} and virtual reality scene reconstruction \cite{serrano2019motion,davidson2020360} have advanced rapidly, 360° panoramic images that are capable of capturing an entire scene in a single frame have increasingly replaced pinhole cameras with limited 60°–120° fields of view as an important medium for scene understanding. Consequently, researchers have turned to semantic segmentation of panoramic images to enhance scene interpretation. However, such images often suffer from severe distortions due to equirectangular projection \cite{yang2021capturing}, making many existing methods, originally developed for pinhole images, ineffective for panoramic segmentation \cite{zhang2022bending}. Beyond distortion, the \textbf{\textit{scarcity of annotated data}} poses a major obstacle to the advancement of panoramic semantic segmentation \cite{cao2025cross}. Creating large-scale annotations \cite{cordts2016cityscapes} is notoriously time-consuming and costly. To address this, researchers have proposed frameworks \cite{zhang2024behind} that transfer knowledge from pinhole image domains, leading to the development of several unsupervised domain adaptation (UDA) techniques \cite{zheng2023both,zhang2024behind}. However, UDA methods typically require access to labeled source-domain data, which may be unavailable in certain scenarios \cite{liu2021source}. To overcome this, recent work \cite{zheng2024360sfuda++} has proposed the \textbf{\textit{Source-Free UDA (SFUDA)}} setting for panoramic segmentation, where only a pre-trained source model (from pinhole images) and unlabeled panoramic images are accessible.

However, self-training domain adaptation methods \cite{zou2019confidence,luo2019taking} rely heavily on high-quality pseudo-labels. This reliance becomes problematic when the source domain has an uneven class distribution \cite{yang2022survey}. The long-tailed distribution \cite{wang2023balancing} leads to under-training on tail classes, further amplifying bias during adaptation and causing significant performance drops for those categories in target domain. To address the challenges of domain shift, data scarcity, and class imbalance in SFUDA panoramic semantic segmentation, this paper proposes a novel framework (\textbf{DAPASS}). As shown in Figure~\ref{fig:Teaser}, DAPASS improves panoramic segmentation performance under adaptation constraints without requiring source data and target labels. Specifically, to combat the issue of unreliable generated pseudo-labels, we designed the \textbf{Panoramic Confidence-Guided Denoising (PCGD)} module. PCGD is able to perform pseudo-label denoising to improve the quality of supervision in the target domain. To achieve robustness against the inherent distortions and scale variations of panoramic images, we introduce the \textbf{Cross-Resolution Attention Module (CRAM)}, which unifies fine-grained local details and wide-angle global context through an attention-driven fusion strategy. The combination boosts the knowledge acuqisition and denoise the knowledge transfer from the source domain to the target domain. We conduct extensive experiments on both outdoor and indoor real-world benchmarks. Our proposed DAPASS consistently outperforms prior state-of-the-art (SOTA) SFUDA methods. Specifically, compared to the previous best-performing method \cite{zheng2024360sfuda++}, DAPASS achieves new SOTA results on the Cityscapes-to-DensePASS (outdoor) and Stanford2D3D (indoor) benchmarks, attaining 55.04\% (\textbf{+2.05\%}) and 70.38\% (\textbf{+1.54\%}) mIoU, respectively, as illustrated in Figure~\ref{fig:Teaser}.

\section{Related Work}
\begin{figure*}[h!]
  \centering
  \includegraphics[width=0.9\textwidth]{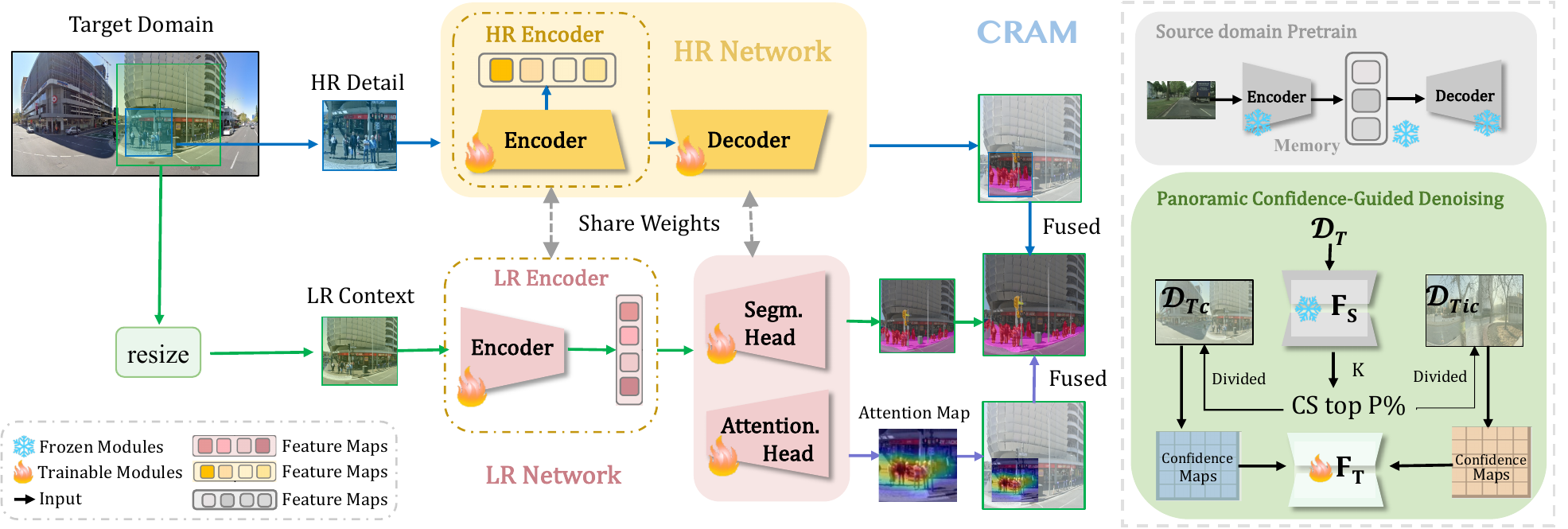}
  \caption{An overview of the proposed DAPASS. The framework consists of two key modules: a Panoramic Confidence-Guided Denoising (PCGD) module to filter noisy pseudo-labels, and a Cross-Resolution Attention Module (CRAM) to enhance detail-level segmentation by combining HR and LR contexts.}
  \vskip -3.0ex
  \label{fig:framework}
\end{figure*}
\subsection{Panoramic Semantic Segmentation}
Panoramic images offer a broader field of view, enabling better scene understanding and enhancing the perception and navigation capabilities of intelligent systems in complex environments \cite{berenguel2023fredsnet,jaus2023panoramic}. To better utilize the advantages of panoramic images, researchers have proposed various methods to obtain more accurate segmentation results. Some researchers have designed attention mechanisms, deformable modules \cite{li2023sgat4pass} and deformable convolution networks \cite{orhan2022semantic,hu2022distortion} to deal with image distortion, the inherent problem in panoramic semantic segmentation. A panel representation that successfully avoids the spatial distortion caused by equirectangular projection (ERP) in the polar regions was proposed \cite{yu2023panelnet}. Because panoramic datasets lack high quality labels, common methods used in pinhole datasets have limited performance in panoramic semantic segmentation.
\subsection{Unsupervised Domain Adaptation (UDA)}
UDA aims to transfer knowledge from a labelled source domain to an unlabelled target domain \cite{araslanov2021self,hoyer2022daformer,liu2022deep,ganin2015unsupervised,long2016unsupervised,sener2016learning, cao2023kt}. To obtain pixel-level segmentation prediction for panoramic images, many studies employ UDA to adapt models trained on pinhole datasets to panoramic datasets \cite{jang2022dada,zhang2022bending,zheng2023look,zheng2023both,ma2021densepass}. For example, Trans4pass \cite{zhang2022bending} proposed a deformable network to handle panoramic distortions and domain shifts. However, most UDA methods require direct access to source data during adaptation, which is impractical in many real-world applications due to data privacy or storage constraints. To deal with such situations, some researchers propose a solution called \textbf{Source-Free Unsupervised Domain Adaptation (SFUDA)} \cite{huang2021model,yeh2021sofa,kundu2021generalize}, which enables adapting a pre-trained source model to the target domain without accessing the source data. To address this constraint, 360SFUDA \cite{zheng2024semantics} was the first framework to introduce SFUDA to the panoramic semantic segmentation task. However, existing methods suffer from a significant decrease in segmentation performance for \textbf{imbalanced categories} and in regions affected by ERP-induced distortions. This degradation is particularly pronounced in 360° panoramic imagery, where the equirectangular projection (ERP) stretches object boundaries and compresses polar areas, leading to spatially varying distortions that do not appear in conventional pinhole images. Moreover, panoramic scenes contain diverse viewing angles and complex global-local interactions, which often result in inconsistent model responses across different regions. To address these challenges, we propose \textbf{DAPASS}, a framework specifically designed to enhance robustness against panoramic distortions while mitigating \textbf{category imbalance} and \textbf{spatial inconsistency} during source-free adaptation.

\vspace{-0.1cm}
\section{Methodology}
\noindent\textbf{Overview.} Given a pre-trained source model $F_S$ and an unlabeled target panoramic domain dataset $\mathcal{D}_T = \{x_t \mid t \in [1, N_t]\}$, where $N_t$ denotes the number of target images, the goal of this SFUDA task is to adapt $F_S$ to $\mathcal{D}_T$, yielding a target model $F_T$ that performs well on panoramic semantic segmentation. To achieve robust segmentation performance in panoramic settings, we propose DAPASS. An overview of the proposed framework is illustrated in Figure~\ref{fig:framework}.
\begin{figure*}[h!]
  \centering
  \includegraphics[width=0.98\textwidth]{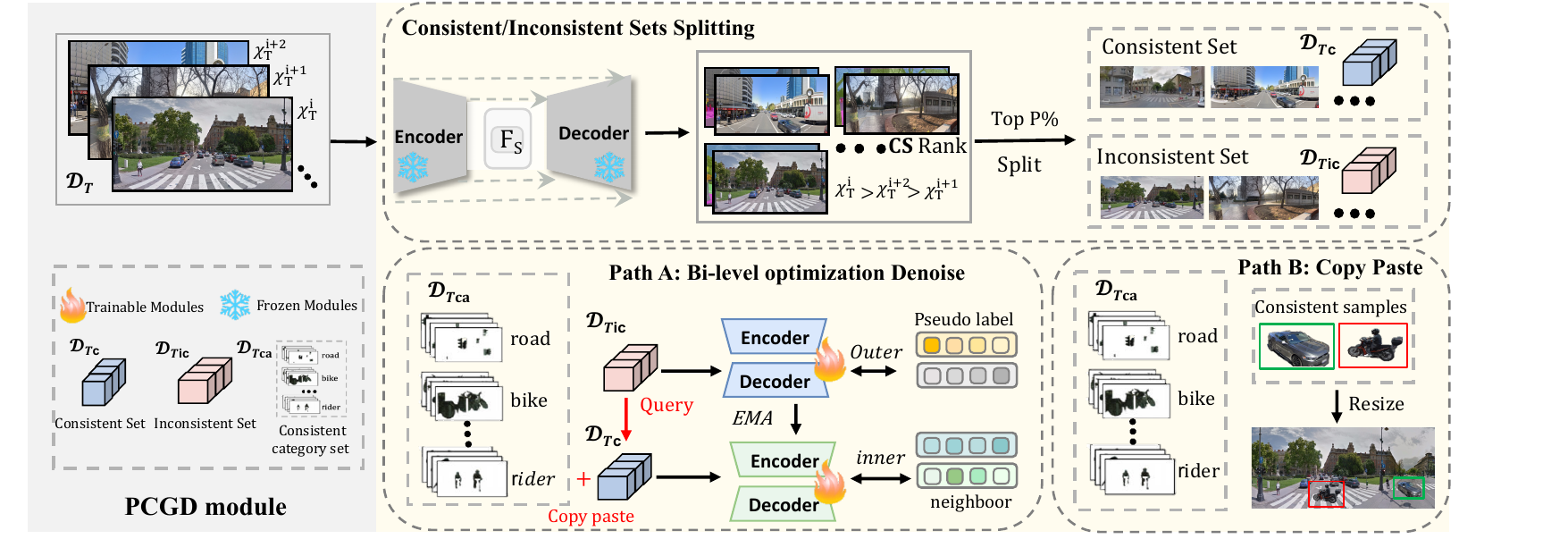}
  \caption{Illustration of the proposed Panoramic Confidence-Guided Denoising (PCGD) module.}
  \label{fig:PCGD}
  \vskip -1.5ex
\end{figure*}
\subsection{Panoramic Confidence-Guided Denoising}
Although the overall DAPASS framework aims to adapt a source-pretrained model to the target panoramic domain, the reliability of pseudo-labels remains critical for stable adaptation. In the source-free setting, pseudo-labels generated by the pretrained model inevitably contain biases: some regions are consistently predicted with high confidence, while others remain uncertain or are systematically ignored. Such uneven supervision hampers stable learning and limits adaptation to diverse target structures. To address this issue, we introduce the \textbf{Panoramic Confidence-Guided Denoising (PCGD)} module, which improves pseudo-label reliability by separating stable and unstable samples via a consistency-based split, leveraging reliable samples to guide noisy ones, and enriching supervision through class-aware copy-paste augmentation, as illustrated in Figure~\ref{fig:PCGD}.

\noindent\textbf{Consistent/Inconsistent Sets Splitting.} 
In UDA tasks, self-training methods~\cite{araslanov2021self, zhao2024stable} observe that samples with stable predictions tend to generalize better, while unstable ones often cause persistent errors due to noise accumulation. This issue becomes more severe in source-free panoramic adaptation, where the domain gap is particularly large and geometric distortion further degrades pseudo-label reliability. 
Therefore, PCGD explicitly divides the target dataset $\mathcal{D}_T$ into \emph{consistent} and \emph{inconsistent} subsets according to a consistency score computed from perturbed predictions of the frozen source model $F_S$. Consistent regions serve as reliable supervision to guide the learning of inconsistent ones, facilitating a more stable and balanced adaptation process.
\begin{equation}
\resizebox{0.85\linewidth}{!}{$
\mathrm{CS}^{i,\tau}
=
-\sum_{l=1}^{H \times W}
D_{\mathrm{KL}}
\!\left(
\mathcal{F}_S(x_t^{(i,l)} \mid \Theta^{0})
\;\middle\|\;
\mathcal{F}_S(x_t^{(i,l)} \mid \Theta^{\tau})
\right)
$}
\label{eq:CS}
\end{equation}
where $D_{\mathrm{KL}}(\cdot \,\|\, \cdot)$ is the Kullback-Leibler (KL) divergence, $x_T^{i,l}$ is the $l$ th pixel of the $i$ th image from the target domain. $\Theta^0$ is the initial parameters of the source model. $\Theta^\tau$ is the parameters of the model after $\tau$ iterations of self-training on the target domain. The higher $CS$, the more consistent samples are. We choose the highest $ \text{top-}P\% $ scores from $CS$ as a basis to divide into consistent sets $\mathcal{D}_{Tc}$ and inconsistent sets $\mathcal{D}_{Tic}$. 

\noindent\textbf{Bi-Level Denoising and Class-Balancing.} After splitting the dataset, two challenges remain: the pseudo-labels in $\mathcal{D}_{Tic}$ are noisy, and the reliable subset $\mathcal{D}_{Tc}$ is still imbalanced and lacks sufficient samples from minority classes (e.g., \textit{Rider}). To address these issues, PCGD adopts a dual-path optimization strategy, as detailed in Algorithm~\ref{alg:pcgd}.

\noindent\textbf{Path A: Bi-Level Neighbor Denoising.} To mitigate pseudo-label noise, we employ a bi-level optimization strategy inspired by~\cite{zhao2024stable}. The key idea is that an update on a noisy sample $x_i \in \mathcal{D}_{Tic}$ is considered reliable only if it also improves performance on a stable sample $x_j \in \mathcal{D}_{Tc}$. Since using the full subsets may suffer from domain or category mismatch, we instead adopt a \textbf{neighbor-guided} strategy: for each $x_i$, we retrieve its most similar stable neighbor $x_j \in \mathcal{D}_{Tc}$ in the feature space and perform bi-level optimization on the pair $(x_i, x_j)$ during adaptation.

\noindent\textbf{Path B: Class Balancing Copy Paste.} While the neighbor-denoising path alleviates pseudo-label noise, it does not explicitly address minority-class scarcity. Existing methods such as SND~\cite{zhao2024stable} may still struggle when reliable samples for rare classes are insufficient. To compensate for this imbalance, we introduce a parallel \textit{balancing path}. Specifically, we first construct a \textbf{Top-$K$ Class-Balancing Pool} $\mathcal{D}_{Tca}^{c}$, which stores high-quality samples for each minority class~$c$. We then apply a copy-paste strategy to provide additional supervision for these rare classes.

\begin{algorithm}[tb]
\caption{Panoramic Confidence-Guided Denoising (PCGD)}
\label{alg:pcgd}
\begin{algorithmic}[1]
\Require Target set $\mathcal{D}_T$, source model $\mathcal{F}_S$, target model $\mathcal{F}_T(\Theta)$
\Ensure Adapted target model $\mathcal{F}_T(\Theta)$

\State \textbf{Consistency-based split:}
\For{$x_t \in \mathcal{D}_T$}
    \State Compute $\mathrm{CS}(x_t)$ using Eq.~\ref{eq:CS}
\EndFor
\State $\mathcal{D}_{Tc} \gets \text{Top-}P\%$ samples ranked by $\mathrm{CS}$, \quad
       $\mathcal{D}_{Tic} \gets \mathcal{D}_T \setminus \mathcal{D}_{Tc}$
\State Build per-class Top-$K$ pool $\mathcal{D}_{Tca}^{c}$ from $\mathcal{D}_{Tc}$

\For{each training iteration}
    \State \textbf{Path A: Neighbor denoising}
    \For{$x_i \sim \mathcal{D}_{Tic}$}
        \State Retrieve nearest stable neighbor $x_j \in \mathcal{D}_{Tc}$
        \State $\Theta_{\text{inner}} \gets \Theta - \alpha \nabla_{\Theta}
        \mathrm{CE}(\mathcal{F}_T(x_i \mid \Theta), \mathcal{F}_S(x_i))$
        \State $\mathcal{L}_{\text{outer}} \gets
        \mathrm{CE}(\mathcal{F}_T(x_j \mid \Theta_{\text{inner}}), \mathcal{F}_S(x_j))$
        \State Update $\Theta$ with $\nabla_{\Theta}\mathcal{L}_{\text{outer}}$
    \EndFor

    \State \textbf{Path B: Class-balancing copy-paste}
    \State Sample $x_p \sim \mathcal{D}_{Tca}^{c}$ and $x_b \sim \mathcal{D}_T$
    \State $(x_{\text{aug}}, y_{\text{aug}}) \gets \Call{CopyPaste}{x_b, x_p}$ using Eq.~\ref{eq:copy_paste}
    \State $\mathcal{L}_{\text{bal}} \gets \mathrm{CE}(\mathcal{F}_T(x_{\text{aug}} \mid \Theta), y_{\text{aug}})$
    \State Update $\Theta$ with $\nabla_{\Theta}\mathcal{L}_{\text{bal}}$
\EndFor
\State \Return $\Theta$
\end{algorithmic}
\end{algorithm}

To execute reliable bi-level optimization, for each inconsistent sample $x_i \in \mathcal{D}_{Tic}$, the goal is to retrieve its most similar counterpart $x_i^{\text{con}} \in \mathcal{D}_{Tc}$ from the consistent set, based on style and layout features. However, the retrieved domain-associated neighbor may still lack certain categories, causing semantic incompleteness. To address this, we propose to borrow the missing class objects from other consistent samples and paste them onto the neighbor for compensation. To enrich the diversity of tail categories, we additionally maintain a Top-$K$ ranked category-specific consistent set $\mathcal{D}_{Tca}^{c}=\{(x^{i},\hat{y}^{i})|x^{i}\in\mathcal{D}_{Tc},CS^{i,m,c}\in \text{Top-$K$}\}$  for each category $c$ as a supplementary resource in $m^{th}$ iteration. Subsequently, the nearest stable neighbor $x_{i}^{con}\in\mathcal{D}_{Tc}$ is augmented using the following copy-paste strategy:
\begin{equation}
(\hat{x}_{i}^{\mathrm{con}}, \hat{y}_{i}^{\mathrm{con}})
= \mathrm{CP}[(x_{i}^{\mathrm{con}}, \hat{y}_{i}^{\mathrm{con}}), (x_{\mathrm{mix}}, \hat{y}_{\mathrm{mix}})]
\label{eq:copy_paste}
\end{equation}
$\text{CP}(\cdot \parallel \cdot)$ denotes the copy-paste operation, which copies objects from the latter to the former. The generated samples are then used in training to further denoise pseudo-labels.
\subsection{Cross-Resolution Attention Module}
ERP panoramas exhibit latitude-dependent distortions, where polar regions are heavily stretched while equatorial regions remain relatively stable. Such geometric inconsistency hinders standard feature learning and becomes more problematic in the source-free setting. To mitigate this issue, CRAM employs two complementary branches: a high-resolution (HR) crop branch for preserving locally reliable details, and a low-resolution (LR) panorama branch for maintaining globally consistent context. Unlike generic multi-resolution fusion, CRAM is designed for panoramic SFUDA, where HR local details are used to calibrate LR global semantics under distortion and scale variation.

\noindent\textbf{LR and HR Crops.}
The combination of multi-resolution inputs enables the model to capture both local details and global scene layout under ERP distortions. The low-resolution context crop $x_t^{LR}$ is obtained by cropping the original high-resolution target image $x_t^{HR}$ and then downsampling the cropped region by a factor of $s$ using bilinear interpolation:
\begin{equation}
x^{LR}_{t}
  = \zeta\!\bigl(\operatorname{Crop}(x_{t}^{HR},\,\mathbf{b}_{L}),\,1/s\bigr),
\label{eq:lr_crop}
\end{equation}
where $\operatorname{Crop}$ denotes the cropping operation, $\mathbf{b}_{L} = (b_{L,1}, b_{L,2}, b_{L,3}, b_{L,4})$ denotes the cropping region, and $\zeta$ denotes bilinear downsampling. We sample the cropping region $\mathbf{b}_{L}$ in three steps. First, we compute the grid interval $k = s \cdot o$, where $o \geq 1$ is the output stride of the segmentation network. We then partition the row and column indices of the high-resolution image into two regular grids with step size $k$, and only grid points are regarded as valid top-left corners for a crop. From each grid, one index is uniformly sampled to obtain the vertical and horizontal start coordinates $(b_{L,1}, b_{L,3})$. To keep the crop inside the image and to fix its size, we set the bottom-right corner as $b_{L,2}=b_{L,1}+s\,h_L$ and $b_{L,4}=b_{L,3}+s\,w_L$. As a result, all crop boundaries lie on the feature-map grid, which guarantees pixel-level correspondence during the subsequent multi-resolution fusion stage.

The high-resolution detail crop is then randomly extracted from the LR context crop:
\begin{equation}
x^{HR}_{t,LR}
  = \operatorname{Crop}(x^{LR}_{t},\,\mathbf{b}_{H}),
\label{eq:hr_crop}
\end{equation}
where $\mathbf{b}_{H} = (b_{H,1}, b_{H,2}, b_{H,3}, b_{H,4})$ denotes the detail cropping region with size $h_H \times w_H$. The context prediction is denoted as $\hat{y}^{LR}_{t} = \mathcal{F}_T^S(x^{LR}_{t})$, and the detail prediction is denoted as $\hat{y}^{HR}_{t,LR} = \mathcal{F}_T^S(x^{HR}_{t,LR})$, where $\mathcal{F}_T^S$ denotes the segmentation network.

\noindent\textbf{Multi-Resolution Fusion.}
To balance fine-grained details with global context, we use a learnable scale-attention mechanism to adaptively fuse predictions from the HR detail crop and the LR context crop. An attention decoder ${F_T}^A$ predicts the scale-attention map $a_{LR} = \sigma\!\left({F_T}^A({F_T}^E(x^{LR}_{t}))\right)$, which weighs the trustworthiness of LR context and HR detail predictions. The sigmoid function $\sigma$ constrains the attention weights to $[0,1]$, where larger values indicate higher reliance on HR detail predictions. The attention is predicted from the LR context crop because it better captures the global scene layout. Since the predictions are smaller than the inputs due to the output stride $o$, the crop coordinates are scaled accordingly in the subsequent fusion steps. Outside the detail crop, the attention is set to zero because no HR detail prediction is available there, yielding the masked attention map $a'_{LR}$. The HR detail prediction is aligned with the upsampled context crop by zero-padding it to a size of $\frac{s h_L}{o} \times \frac{s w_L}{o}$, resulting in the aligned HR prediction $\hat{y}_{LR}^{HR}$. The final fused prediction is computed as
\begin{equation}
\hat{y}_{LR,F}
=
\zeta((1-a'_{LR}) \odot \hat{y}_{LR}, s)
+
\zeta(a'_{LR}, s) \odot \hat{y}_{LR}^{HR}.
\label{eq:scale_fusion}
\end{equation}

The encoder, segmentation head, and attention head are jointly trained using both the fused multi-resolution prediction and the HR detail prediction. The target loss $\mathcal{L}_{\mathrm{CRAM}}^{T}$ is defined as
\begin{equation}
\begin{aligned}
\mathcal{L}_{\mathrm{CRAM}}^{T}
  &= (1-\lambda_d)\,
     \mathcal{L}_{\mathrm{ce}}\!\bigl(\hat{y}_{LR,F}^{T},\,p_{LR,F}^{T},\,q_{LR,F}^{T}\bigr) \\[2pt]
  &\quad + \lambda_d\,
     \mathcal{L}_{\mathrm{ce}}\!\bigl(\hat{y}_{HR}^{T},\,p_{HR}^{T},\,q_{HR}^{T}\bigr),
\end{aligned}
\label{eq:l_cram_target}
\end{equation}
where $\lambda_d \in [0,1]$ is the balance factor between the low-resolution context branch and the high-resolution detail branch. $\mathcal{L}_{\mathrm{ce}}(\cdot,\cdot,\tau)$ denotes the temperature-scaled cross-entropy, where temperature $\tau$ sharpens ($\tau<1$) or smooths ($\tau>1$) the pseudo-label distribution. $\hat{y}^{T}_{LR,F}$ denotes the logits of the fused context prediction after upsampling to the original image size, and $p^{T}_{LR,F}$ denotes the corresponding pseudo-label map. $\hat{y}^{T}_{HR}$ denotes the logits predicted on the high-resolution detail crop, and $p^{T}_{HR}$ denotes the corresponding pseudo-labels for the detail crop.
\begin{figure*}[t]
  \centering
  \includegraphics[height=7cm]{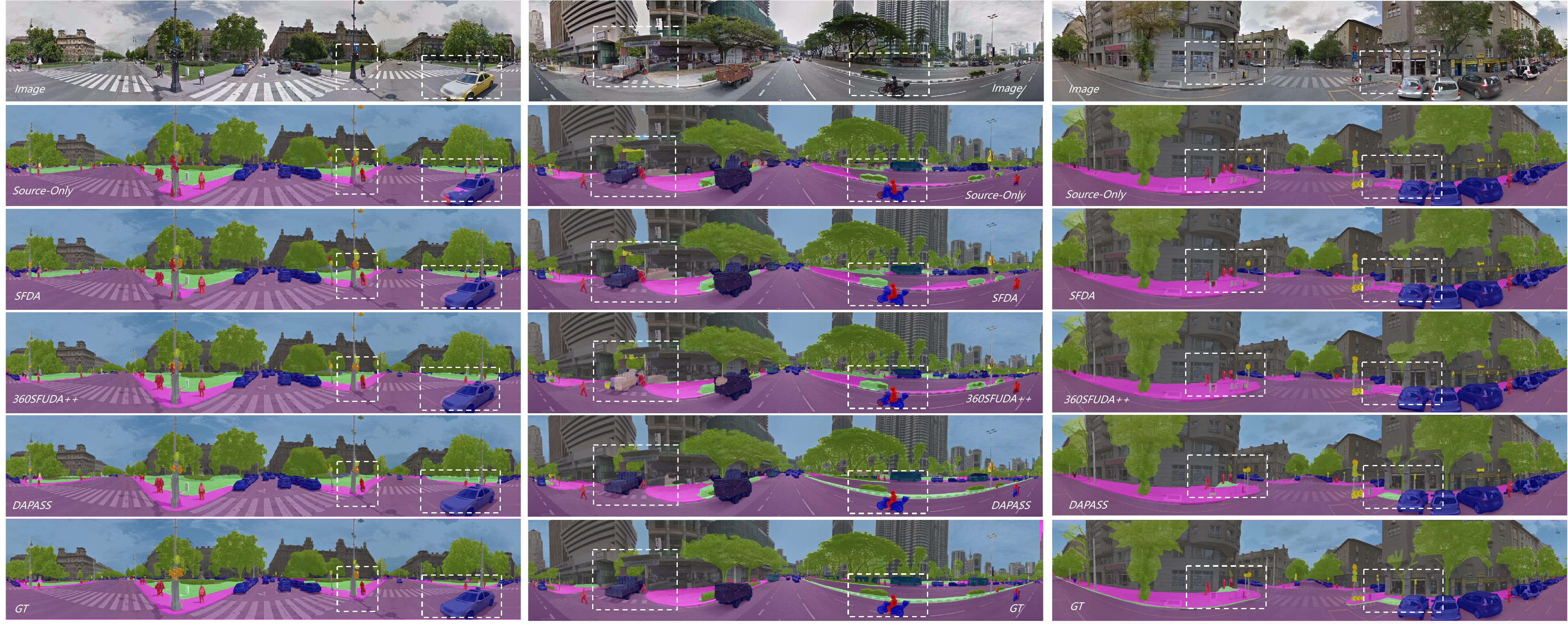}
  \vskip-1.5ex
  \caption{\textbf{Visualization results on C-to-D scenario}. From top to bottom: input image, Source-Only, SFDA, 360SFUDA++ \cite{zheng2024360sfuda++}, our DAPASS, and ground truth (GT). Compared to prior methods, DAPASS yields more accurate boundaries, reduces noise, and better recognizes occluded objects (highlighted in dashed boxes).
}
  \label{fig:C2D results}
\end{figure*}
\begin{table*}[t]
\centering
\renewcommand{\tabcolsep}{1.5pt}
\resizebox{\textwidth}{!}{
\begin{tabular}{l|c|c|cccccccccccccccccccc}
\toprule
Method & SF & mIoU & Road & S.W. & Build. & Wall & Fence & Pole & Light & Sign & Vegt. & Terr. & Sky & Pers. & Rider & Car & Truck & Bus & Train & Motor & Bike \\
\midrule
ECANet \cite{yang2021capturing} & $\times$ & 43.02 & 81.60 & 19.46 & 81.00 & 32.02 & 39.47 & 25.54 & 3.85 & 17.38 & 79.01 & 39.75 & 94.60 & 46.39 & 12.98 & 81.96 & 49.25 & 28.29 & 0.00 & 55.36 & 29.47 \\
P2PDA \cite{zhang2021transfer} & $\times$ & 41.99 & 70.21 & 30.24 & 78.44 & 26.72 & 28.44 & 14.02 & 11.67 & 5.79 & 68.54 & 38.20 & 85.97 & 28.14 & 0.00 & 70.36 & 60.49 & 38.90 & 77.80 & 39.85 & 24.02 \\
SIM\cite{wang2020differential} & $\times$ & 44.58 & 68.16 & 32.59 & 80.58 & 25.68 & 31.38 & 23.60 & 19.39 & 14.09 & 72.65 & 26.41 & 87.88 & 41.74 & 16.09 & 73.56 & 47.08 & 42.81 & 56.35 & 47.72 & 39.30 \\
PCS\cite{yue2021prototypical} & $\times$ & 53.83 & 78.10 & 46.24 & 86.24 & 30.33 & 45.78 & 34.04 & 22.74 & 13.00 & 79.98 & 33.07 & 93.44 & 47.69 & 22.53 & 79.20 & 61.59 & 67.09 & 83.26 & 58.68 & 48.13 \\
DAFormer \cite{hoyer2022daformer} & $\times$ & 54.67 & 73.75 & 27.34 & 86.35 & 35.88 & 45.56 & 36.28 & 25.53 & 10.65 & 79.87 & 41.64 & 94.74 & 49.69 & 25.15 & 77.70 & 63.06 & 65.61 & 86.68 & 65.12 & 48.13 \\
Trans4PASS-T \cite{zhang2022bending} & $\times$ & 53.18 & 78.13 & 41.19 & 85.93 & 29.88 & 37.02 & 32.54 & 21.59 & 18.94 & 78.67 & 45.20 & 93.88 & 48.54 & 16.91 & 79.58 & 65.33 & 55.76 & 84.63 & 59.05 & 37.61 \\
DPPASS-T \cite{zheng2023both} & $\times$ & 55.30 & 78.74 & 46.29 & 87.47 & 48.62 & 40.47 & 35.38 & 24.97 & 17.39 & 79.23 & 40.85 & 93.49 & 52.09 & 29.40 & 79.19 & 58.73 & 47.24 & 86.48 & 66.60 & 38.11 \\
DATR-S \cite{zheng2023look} & $\times$ & 56.81 & 80.63 & 51.77 & 87.80 & 44.94 & 43.73 & 37.23 & 25.66 & 21.00 & 78.61 & 26.68 & 93.77 & 54.62 & 29.50 & 80.03 & 67.35 & 63.75 & 87.67 & 67.57 & 37.10 \\
\arrayrulecolor{black}\specialrule{.1em}{.05em}{.05em} 
Source-Only & \checkmark & 38.65 & 65.26 & 29.40 & 77.04 & 15.14 & 28.72 & 14.15 & 9.36 & 10.55 & 69.09 & 21.10 & 82.91 & 40.98 & 10.42 & 68.56 & 32.90 & 44.94 & 50.98 & 37.74 & 25.19 \\
SFDA \cite{liu2021source} & \checkmark & 42.70 & 68.75 & 31.59 & 80.99 & 19.61 & 29.60 & 18.67 & 7.7 & 14.08 & 73.74 & 24.91 & 88.38 & 41.66 & 8.46 & 69.97 & 47.48 & 33.25 & 72.02 & 47.62 & 32.77\\
DATC \cite{yang2022source} & \checkmark & 43.06 & 70.21 & 35.87 & 80.60 & 21.42 & 28.14 & 19.10 & 5.79 & 15.10 & 72.76 & 27.42 & 88.14 & 41.65 & 10.29 & 72.32 & 47.80 & 21.97 & 80.91 & 46.65 & 32.01 \\
360SFUDA++ \cite{zheng2024360sfuda++} w/ b1 & \checkmark & 50.19 & 68.59 & 46.70 & 84.06 & 29.08 & 34.41 & 31.28 & 20.31 & 16.84 & 75.04 & 23.07 & 92.20 & 50.03 & 18.32 & 78.75 & 56.53 & 49.90 & 83.63 & 59.00 & 35.46 \\
360SFUDA++ \cite{zheng2024360sfuda++} w/ b2 & \checkmark & 52.99 & 74.00 & \textbf{48.03} & 85.86 & \textbf{34.41} & 41.28 & 31.25 & \textbf{22.59} & 17.41 & 76.74 & 26.24 & 92.65 & \textbf{54.60} & \textbf{30.76} & 79.41 & 55.60 & 59.84 & 81.25 & 59.89 & 35.02 \\
\rowcolor{gray!10} Ours DAPASS (w/ b1) & \checkmark & 53.16 & 78.15 & 41.51 & 85.74 & 29.48 & 37.46 & 32.24 & 21.89 & \textbf{19.14} & 78.82 & \textbf{45.38} & \textbf{93.73} & 49.02 & 16.74 & 79.32 & 64.84 & 56.02 & \textbf{84.52} & 58.58 & 37.43 \\
\rowcolor{gray!10} Ours DAPASS (w/ b2) & \checkmark & \textbf{55.04} & \textbf{78.75} & 41.95 & \textbf{86.28} & 31.89 & \textbf{44.79} & \textbf{34.32} & 22.51 & 17.22 & \textbf{79.47} & 42.11 & 93.48 & 49.79 & 20.97 & \textbf{81.34} & \textbf{66.84} & \textbf{68.77} & 84.39 & \textbf{61.23} & \textbf{39.74} \\
\bottomrule
\end{tabular}
}
\vskip-1.5ex
\caption{Experimental results of 19 classes in panoramic semantic segmentation on the C-to-D setting. SF: source-free UDA. Bold values indicate the best performance among source-free methods.}
\label{tab:sfuda_benchmark}
\vskip-3ex
\end{table*}
\section{Experiments}
\subsection{Experiment Setup}
\noindent\textbf{Datasets.} We validate our method on three benchmark datasets that encompass outdoor and indoor scenarios. \textbf{Cityscapes} \cite{cordts2016cityscapes} is a real-world dataset featuring urban street scenes, widely adopted for autonomous driving research. It offers precise pixel-wise annotations across 19 semantic categories. The dataset’s official split includes 2975 images for training and 500 images for validation. In this paper, we utilise the official training set (2975 images) as the source domain data for pre-training the source model. \textbf{DensePASS} \cite{ma2021densepass} is a panoramic dataset designed to capture diverse street-level perspectives and contains 2,500 panoramic images in total. Of these, 100 panoramas are meticulously annotated to cover categories that are especially critical for navigation; this annotated subset serves as the test set during evaluation. In the C-to-D setting, we regard the training portion of DensePASS as unlabeled target-domain panoramic data. \textbf{Stanford2D3D} \cite{armeni2017joint} contains 70,496 pinhole-view images labeled with 13 semantic classes. In contrast, the Stanford 2D-3D Panoramic (SPan) dataset offers 1,413 indoor panoramas, each annotated with the same 13 classes as SPin. For our experiments, we concentrate only on the eight semantic classes that appear in both datasets. Accordingly, our experiments are conducted under both real-world adaptation settings (Cityscapes-to-DensePASS, abbreviated as \textbf{C-to-D}, and Stanford2D3D-pinhole-to-Stanford2D3D-panoramic, \textbf{SPin-to-SPan}). This rigorous setup allows for a thorough assessment of our model's effectiveness and robustness across varied conditions.

\noindent\textbf{Minority/Majority Class Definition.}
For class-group analysis, minority classes are defined by low occurrence frequency in the target domain. In DensePASS (C-to-D), the minority classes are \textit{Pole, Light, Sign, Person, Rider, Truck, Bus, Train, Motor,} and \textit{Bike}. In SPin-to-SPan, the minority classes are \textit{Chair, Door, Sofa, Table,} and \textit{Window}. The remaining classes are treated as majority classes.
\begin{figure*}[t]
  \centering
  \includegraphics[width=\textwidth]{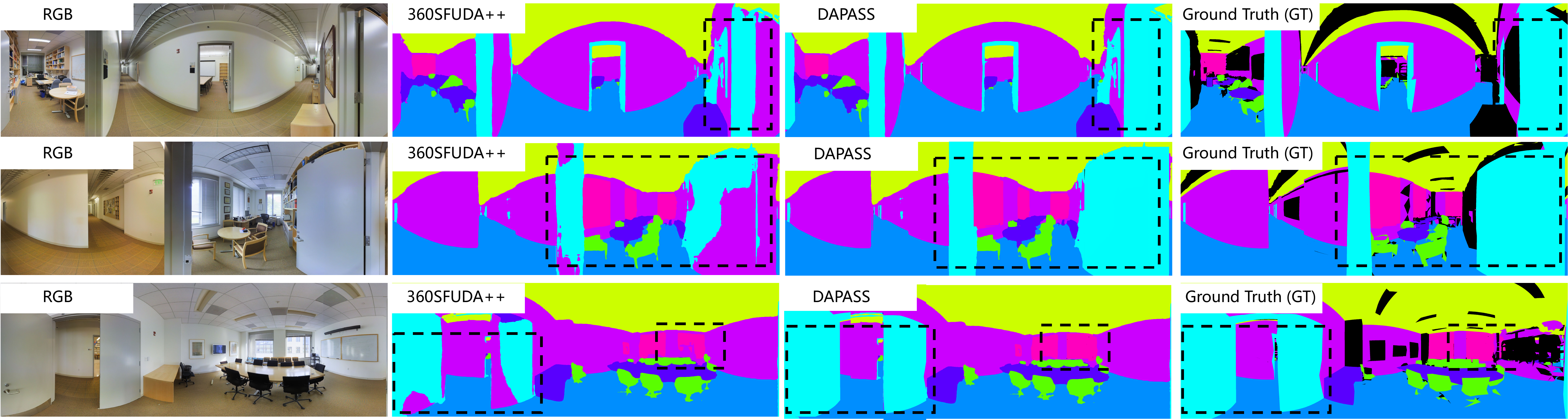}
  \caption{Visualization results on Spin-to-Span task. From left to right: input image, 360SFUDA++ \cite{zheng2024360sfuda++}, our DAPASS, and GT.}
  \label{fig: SPin-to-SPan SFUDA results}
  \vskip-1.0ex
\end{figure*}
\begin{table*}[h!]
\centering
\renewcommand{\tabcolsep}{8pt}
\resizebox{\textwidth}{!}{
\begin{tabular}{l|c|c|cccccccccc}
\toprule
Method & SF & mIoU & Ceiling & Chair & Door & Floor & Sofa & Table & Wall & Window & $\Delta$ \\
\midrule
PVT-S \cite{zhang2022bending}& $\times$ & 57.71 & 85.69 & 51.71 & 18.54 & 90.78 & 34.76 & 65.34 & 74.87 & 39.98 & - \\
PVT-S w/ MPA \cite{zhang2022bending}& $\times$ & 57.95 & 85.85 & 51.76 & 18.39 & 90.78 & 35.93 & 65.43 & 75.00 & 40.43 & - \\
Trans4PASS w/ MPA \cite{zhang2022bending} & $\times$ & 64.52 & 85.08 & 58.72 & 34.97 & 91.12 & 46.25 & 71.72 & 77.58 & 50.75 & - \\
Trans4PASS+ \cite{zhang2022bending} & $\times$ & 63.73 & 90.63 & 62.30 & 24.79 & 92.62 & 35.73 & 73.16 & 78.74 & 51.78 & - \\
Trans4PASS+ w/ MPA \cite{zhang2022bending} & $\times$ & 67.16 & 90.04 & 64.04 & 42.89 & 91.74 & 38.34 & 71.45 & 81.24 & 57.54 & - \\
\arrayrulecolor{black}\specialrule{.1em}{.05em}{.05em} 
SFDA \cite{liu2021source} & \checkmark & 54.76 & 79.44 & 33.20 & 52.09 & 67.36 & 22.54 & 53.64 & 69.38 & 60.46 & - \\
360SFUDA++ w/ b1 \cite{zheng2024360sfuda++} & \checkmark & 66.54 & 86.28 & 58.25 & 36.04 & 87.99 & 48.74 & 65.44 & 77.82 & \textbf{71.78} & - \\
360SFUDA++ w/ b2 \cite{zheng2024360sfuda++} & \checkmark & 68.84 & 85.50 & 57.59 & 53.15 & 87.40 & \textbf{53.63} & 66.49 & 80.23 & 66.75 & - \\
\rowcolor{gray!10} Ours DAPASS (w/ b1) & \checkmark & 69.56 & 89.64 & \textbf{65.50}& 45.83 & 94.16 & 47.07 & \textbf{75.96} & 81.93 & 56.39 & +14.80 \\
\rowcolor{gray!10} Ours DAPASS (w/ b2)& \checkmark & \textbf{70.38} & \textbf{89.59} & 64.81 & \textbf{53.33} & \textbf{94.55} & 46.76 & 73.90 & \textbf{83.03} & 57.08 & \textbf{+15.62} \\
\bottomrule
\end{tabular}
}
\caption{Experimental results of 8 classes on Spin-to-Span setting. Bold values indicate best performance among SF methods. $\Delta$ is an improvement over SFDA \cite{liu2021source}.}
\label{tab:indoor_sfuda_comparison}
\vskip-3ex
\end{table*}

\noindent\textbf{Implementation Details.} We train our models with 4 3090 GPUs with an initial learning rate of $6 {\times} 10^{-5}$, which is scheduled by the poly strategy with power 0.9 over 200 epochs. The optimizer is AdamW with epsilon $= 10^{-8}$, weight decay $10^{-4}$, and a batch size of~4 on each GPU.  
The adaptation frameworks are trained for $12\,\text{K}$ iterations. Training images captured by Stanford2D3D pinhole cameras and panoramic views are processed at resolutions of \(1080\times1080\) and \(1024\times512\), respectively.
For outdoor scenarios, Cityscapes pinhole images are adjusted to \(1024\times512\), whereas DensePASS panoramic images are presented at \(2048\times400\).  To ensure uniformity in the evaluation framework, the resolutions for validation image sets are fixed at \(2048\times1024\) for indoor environments and \(2048\times400\) for outdoor environments. In Spin-to-Span task, we selected the hyperparameters top-$P\%$, $\tau$ and $s$ as 15.0, 8,000 and 2, respectively.  
In C-to-D setting, we selected the hyperparameters top-$P\%$, $\tau$ and $s$ as 10.0, 6,000 and 2, respectively. 
\subsection{Experiment Results}
In this section, we compare DAPASS with previous SOTA approaches in different settings. Firstly, we undertake a comprehensive evaluation of the proposed DAPASS within the C-to-D scenario, including both qualitative and quantitative analyses. As shown in Figure \ref{fig:C2D results} and Table \ref{tab:sfuda_benchmark}, DAPASS establishes new SOTA performance on outdoor C-to-D task,  demonstrating the adaptation capabilities across urban panoramas. DAPASS surpasses the previous SOTA, 360SFUDA++ \cite{zheng2024360sfuda++}, with mIoU increases of 2.97\% and 2.05\% using SegFormer-B1 and -B2 backbones. And it significantly outperforms SFUDA methods SFDA \cite{liu2021source} and DATC \cite{yang2022source} by 10.46\% and 10.10\% in mIoU, respectively. The qualitative results illustrate that DAPASS is capable of producing finer boundary segmentation, accurately recognizing small objects, and maintaining better semantic consistency with the ground truth, significantly outperforming other compared methods. Subsequently, we conduct an evaluation of our DAPASS within the Spin-to-Span scenario. The outcomes of this assessment are detailed in Table \ref{tab:indoor_sfuda_comparison} and Figure \ref{fig: SPin-to-SPan SFUDA results}. We compare it with previous SOTA 360SFUDA++ \cite{zheng2024360sfuda++}, SFUDA methods and the UDA methods. As shown in Table \ref{tab:indoor_sfuda_comparison}, DAPASS outperforms 360SFUDA++ \cite{zheng2024360sfuda++} and SFDA \cite{liu2021source} by 3.02\% and 14.80\%, respectively. Notably, DAPASS surpasses the UDA methods MPA \cite{zhang2022bending} that is accessible to source data in training process by 3.22\%. The visualization results in Figure \ref{fig: SPin-to-SPan SFUDA results} demonstrate that DAPASS is able to deal with the severe distortions and structural complexities commonly found in panoramic indoor scenes, producing more accurate and consistent segmentation results compared to 360SFUDA++ \cite{zheng2024360sfuda++}. \textbf{\textit{More quantitative comparison and visualization results refer to the supplementary material}}.
\section{Ablation Study}
We conduct a series of ablation studies on the both \textbf{indoor (SPin-to-SPan)} and \textbf{outdoor (C-to-D) scenarios} to evalute the effectiveness of our method.
\begin{table}
    \centering
    \small
    \begin{tabular}{l|cc|cc}
        \toprule
        \multirow{2}{*}{Method} & \multicolumn{2}{c|}{C-to-D} & \multicolumn{2}{c}{Spin-to-Span} \\
         & B1 & B2 & B1 & B2 \\
        \midrule
        Source-Only & 36.43 & 38.65 & 43.54 & 46.75 \\
        Unweighted Pseudo-Labels & 40.71 & 42.56 & 51.35 & 53.54 \\
        w/o Path A  & 47.82 & 49.10 & 61.43 & 63.55 \\
        w/o Path B  & 48.35 & 50.02 & 63.12 & 64.78 \\
        \underline{PCGD (Full, ours)} & \textbf{50.23} & \textbf{52.38} & \textbf{65.87} & \textbf{67.32} \\
        \bottomrule
    \end{tabular}
    \caption{Ablation study of \textbf{PCGD}. Path A: bi-level neighbor denoising. Path B: class-balancing copy–paste. B1/B2 denote two SegFormer backbones. Metric: mIoU.}
    \label{tab:effectivePCGD}
    \vskip-3ex
\end{table}

\noindent\textbf{Effectiveness of PCGD.} Ablation experiments were conducted to analyze the individual contributions of the two components in PCGD. As shown in Table~\ref{tab:effectivePCGD}, directly relying on the source model or training with unweighted pseudo-labels leads to limited improvements under the source-free setting.
Removing \textbf{Path A} (bi-level neighbor denoising) causes a clear performance drop across all settings (e.g., from 50.23~$\rightarrow$~47.82 mIoU on C-to-D B1), confirming that Path A plays a central role in stabilizing pseudo-labels and suppressing noisy updates. Removing \textbf{Path B} (class-balancing copy--paste) also results in degraded performance (50.23~$\rightarrow$~48.35 on C-to-D B1), particularly on Spin-to-Span where tail-class imbalance is more severe. This verifies that Path B effectively complements Path A by enriching minority-class samples and improving class coverage. When both paths are enabled, PCGD achieves the best results, outperforming all variants on both datasets and backbones. These findings demonstrate that Path A and Path B address complementary aspects of pseudo-label quality—noise reduction and class balancing—and that their synergy is essential for robust SFUDA adaptation.
 
\begin{table}
    \centering
    \small
    \begin{tabular}{l|cc|cc}
        \toprule
        \multirow{2}{*}{Method} & \multicolumn{2}{c|}{C-to-D} & \multicolumn{2}{c}{Spin-to-Span} \\
        & B1 & B2 & B1 & B2 \\
        \hline
        \midrule
        Source-Only & 36.43 & 38.65 & 43.54 & 46.75 \\
        SFDA & 42.70 & 44.10 & 54.76 & 56.32 \\
        PCGD (only) & 50.23 & 52.38 & 65.87 & 67.32 \\
        \underline{PCGD + CRAM (ours)} & 53.16 & 55.04 & 69.56 & 70.38 \\
        \bottomrule
    \end{tabular}
    \caption{Ablation analysis for \textbf{CRAM}. B1 and B2 denote two different SegFormer backbones \cite{xie2021segformer}. Evaluation metric is mIoU.}
    \label{tab:effectiveCRAM}
    \vskip-1ex
\end{table}

\noindent\textbf{Effectiveness of CRAM.} To assess the contribution of the proposed CRAM module, we conduct an ablation study by comparing the performance of the model with and without CRAM under the same PCGD framework. As shown in Table \ref{tab:effectiveCRAM}, adding CRAM consistently improves the mIoU across all settings. For instance, on the C-to-D task, CRAM brings an additional gain of 2.93\% and 2.66\% on B1 and B2 backbones, respectively. Similarly, for the Spin-to-Span task, CRAM improves the mIoU by 3.69\% (B1) and 3.06\% (B2). These consistent improvements demonstrate that CRAM complements PCGD by enhancing cross-domain representation alignment and reinforcing spatial consistency during adaptation.
\begin{table}[tbp]
  \centering
   \renewcommand{\arraystretch}{1.05}
  \resizebox{0.42\textwidth}{!}{
    \begin{tabular}{c|ccccc}
    \toprule
    $P $ (\%) & 1.0  & 5.0 & 10.0 & 15.0 & 20.0 \\
    \midrule
    C-to-D  & 54.52 & 54.76  & \textbf{55.04}  & 54.86  & 54.91 \\
    Spin-to-Span   & 69.35  & 69.67  & 69.88  &  \textbf{70.38}  & 69.92 \\
    \midrule
    \midrule
    $\tau$ $(\times 1000)$ & 2     & 4     & 6     & 8    & 10 \\
    \midrule
    C-to-D      & 54.23  & 54.65  & \textbf{55.04}  & 55.04  & 55.04  \\
    Spin-to-Span  & 69.35  & 69.77  & 69.92 & \textbf{70.38}  & 70.38 \\
    \bottomrule
    \end{tabular}%
     }
  \caption{Sensitivity study of the hyper-parameter $P$ and $\tau$. All experiments use SegFormer‑B2\cite{xie2021segformer} as the backbone.}
  \label{tab:Hyper}%
  \vskip-3ex
\end{table}%

\noindent\textbf{Hyperparameter Sensitivity.}
We analyze the sensitivity of the hyperparameters $P$ and $\tau$ in Eq.~\ref{eq:CS} under both C-to-D and SPin-to-SPan settings. As shown in Table~\ref{tab:Hyper}, for the top-$P\%$ selection ratio, we vary $P$ from 1 to 20. A very small $P$ may fail to capture sufficient informative samples, while a large $P$ risks introducing noisy or unreliable samples into the consistent set. Across both tasks, the mIoU fluctuations remain within a small range, indicating that PCGD is relatively insensitive to the choice of $P$. We further vary $\tau$, the number of iterations used for consistency-score computation, from 2,000 to 10,000. The results show that different choices of $\tau$ lead to only marginal changes, suggesting that the consistent/inconsistent split is stable. Additional analysis further shows that the copy-paste pool size $K \in [10,20]$ yields stable performance, the loss weight $\lambda_d$ remains robust within $[0.1,0.5]$, and the similarity metric used for Path A neighbor retrieval (Cosine vs.\ L2) results in negligible differences. Detailed results are provided in the supplementary material.

\vspace{-0.1cm}
\section{Discussion}
Although our class-balancing copy--paste module improves minority-class supervision, the results in 
Table~\ref{tab:sfuda_benchmark} and Table~\ref{tab:indoor_sfuda_comparison} show that several tail classes (e.g., \textit{Rider}, 
\textit{Person}) still exhibit larger performance fluctuations. This is partly because the module relies on 
the quality of pseudo-labels for rare categories; when such classes appear infrequently or have unstable 
masks, the constructed Top-$K$ pool may contain imperfect samples.

In addition, the current copy--paste strategy is relatively naive: without contextual or geometric 
alignment, pasted objects may appear in unrealistic locations and introduce noise. Nevertheless, our 
results suggest that its overall benefit still dominates, and moderate contextual disruption may even 
act as a regularizer by reducing background bias and encouraging stronger object cues. Future work 
could explore geometry-aware pasting or diffusion-based blending for more reliable minority-class modeling.

\vspace{-0.1cm}
\section{Conclusion}
In this paper, we explored the challenging problem of source-free unsupervised domain adaptation from pinhole to panoramic domain. To effectively address cross-domain semantic segmentation difficulties under this setting, we propose a novel framework named DAPASS. DAPASS tackles several domain adaptation challenges, including semantic misalignment, inherent distortions in panoramic images, and the noisy pseudo-labels in self-training. DAPASS enables robust cross-domain knowledge transfer without access to source data. To be specific, PCGD generates high-quality and class-balanced pseudo-labels, enhancing the accuracy of supervision signals during self-training and facilitating efficient knowledge transfer. Meanwhile, CRAM integrates high-resolution details and low-resolution global context, allowing the model to better focus on image details under distortion and thus achieve more reliable knowledge adaptation.
Extensive experiments on indoor and outdoor benchmarks demonstrate that DAPASS outperforms existing SFUDA methods and remains competitive with source-dependent UDA approaches.
\clearpage
\section*{Acknowledgements} This research was jointly supported by the National Natural Science Foundation of China Projects (Grants No.~42401543) and the National Natural Science Foundation of China (General Program) (Grant No.~42571521).
{
    \small
    \bibliographystyle{unsrtnat}
    \bibliography{main}

@String(CVPR= {IEEE Conf. Comput. Vis. Pattern Recog.})

@String(ICCV= {Int. Conf. Comput. Vis.})

@String(ICME = {Int. Conf. Multimedia and Expo})

@String(AAAI = {AAAI})

@String(CVPR  = {CVPR})

@String(ICCV  = {ICCV})

@String(ICME  =	{ICME})

@inproceedings{ma2021densepass,
  author    = {Ma, Chaoxiang and Zhang, Jiaming and Yang, Kailun and Roitberg, Alina and Stiefelhagen, Rainer},
  title     = {DensePASS: Dense Panoramic Semantic Segmentation via Unsupervised Domain Adaptation with Attention-Augmented Context Exchange},
  booktitle = {Proceedings of the IEEE International Conference on Intelligent Transportation Systems (ITSC)},
  pages     = {2766--2772},
  year      = {2021},
  publisher = {IEEE}
}

@inproceedings{davidson2020360,
  title={360 camera alignment via segmentation},
  author={Davidson, Benjamin and Alvi, Mohsan S and Henriques, Jo{\~a}o F},
  booktitle={European conference on computer vision},
  pages={579--595},
  year={2020},
  organization={Springer}
}

@article{serrano2019motion,
  author    = {Serrano, Ana and Kim, Incheol and Chen, Zhili and DiVerdi, Stephen and Gutierrez, Diego and Hertzmann, Aaron and Masia, Belen},
  title     = {Motion Parallax for 360 RGBD Video},
  journal   = {IEEE Transactions on Visualization and Computer Graphics},
  volume    = {25},
  number    = {5},
  pages     = {1817--1827},
  year      = {2019},
  publisher = {IEEE}
}

@article{yang2022survey,
  title={A survey on long-tailed visual recognition},
  author={Yang, Lu and Jiang, He and Song, Qing and Guo, Jun},
  journal={International Journal of Computer Vision},
  volume={130},
  number={7},
  pages={1837--1872},
  year={2022},
  publisher={Springer}
}

@inproceedings{liu2021source,
  author    = {Liu, Yuang and Zhang, Wei and Wang, Jun},
  title     = {Source-Free Domain Adaptation for Semantic Segmentation},
  booktitle = {Proceedings of the IEEE/CVF Conference on Computer Vision and Pattern Recognition (CVPR)},
  pages     = {1215--1224},
  year      = {2021},
  publisher = {IEEE}
}

@inproceedings{berenguel2023fredsnet,
  title={Fredsnet: Joint monocular depth and semantic segmentation with fast fourier convolutions from single panoramas},
  author={Berenguel-Baeta, Bruno and Bermudez-Cameo, Jesus and Guerrero, Jose J},
  booktitle={2023 IEEE International Conference on Robotics and Automation (ICRA)},
  pages={6080--6086},
  year={2023},
  organization={IEEE}
}

@article{jaus2023panoramic,
  author    = {Jaus, Alexander and Yang, Kailun and Stiefelhagen, Rainer},
  title     = {Panoramic Panoptic Segmentation: Insights into Surrounding Parsing for Mobile Agents via Unsupervised Contrastive Learning},
  journal   = {IEEE Transactions on Intelligent Transportation Systems},
  volume    = {24},
  number    = {4},
  pages     = {4438--4453},
  year      = {2023},
  publisher = {IEEE}
}

@article{orhan2022semantic,
  author    = {Semih Orhan and Yalin Bastanlar},
  title     = {Semantic Segmentation of Outdoor Panoramic Images},
  journal   = {Signal, Image and Video Processing},
  volume    = {16},
  number    = {3},
  pages     = {643--650},
  year      = {2022},
  publisher = {Springer}
}

@article{hu2022distortion,
  author    = {Xing Hu and Yi An and Cheng Shao and Huosheng Hu},
  title     = {Distortion Convolution Module for Semantic Segmentation of Panoramic Images Based on the Image-Forming Principle},
  journal   = {IEEE Transactions on Instrumentation and Measurement},
  volume    = {71},
  pages     = {1--12},
  year      = {2022},
  publisher = {IEEE}
}

@article{li2023sgat4pass,
  author    = {Xuewei Li and Tao Wu and Zhongang Qi and Gaoang Wang and Ying Shan and Xi Li},
  title     = {{SGAT4PASS}: Spherical Geometry-Aware Transformer for Panoramic Semantic Segmentation},
  journal   = {arXiv preprint arXiv:2306.03403},
  year      = {2023}
}

@inproceedings{yu2023panelnet,
  author    = {Haozheng Yu and Lu He and Bing Jian and Weiwei Feng and Shan Liu},
  title     = {PanelNet: Understanding 360 Indoor Environment via Panel Representation},
  booktitle = {Proceedings of the IEEE/CVF Conference on Computer Vision and Pattern Recognition (CVPR)},
  pages     = {878--887},
  year      = {2023},
  publisher = {IEEE}
}

@article{jang2022dada,
  title={DaDA: Distortion-aware domain adaptation for unsupervised semantic segmentation},
  author={Jang, Sujin and Na, Joohan and Oh, Dokwan},
  journal={Advances in Neural Information Processing Systems},
  volume={35},
  pages={18681--18693},
  year={2022}
}

@inproceedings{zhang2022bending,
  author    = {Jiaming Zhang and Kailun Yang and Chaoxiang Ma and Simon Rei{\ss} and Kunyu Peng and Rainer Stiefelhagen},
  title     = {Bending Reality: Distortion-Aware Transformers for Adapting to Panoramic Semantic Segmentation},
  booktitle = {Proceedings of the IEEE/CVF Conference on Computer Vision and Pattern Recognition (CVPR)},
  pages     = {16917--16927},
  year      = {2022},
  publisher = {IEEE}
}

@inproceedings{zheng2023both,
  author    = {Xu Zheng and Jinjing Zhu and Yexin Liu and Zidong Cao and Chong Fu and Lin Wang},
  title     = {Both Style and Distortion Matter: Dual-Path Unsupervised Domain Adaptation for Panoramic Semantic Segmentation},
  booktitle = {Proceedings of the IEEE/CVF Conference on Computer Vision and Pattern Recognition (CVPR)},
  pages     = {1285--1295},
  year      = {2023},
  publisher = {IEEE}
}

@inproceedings{araslanov2021self,
  author    = {Nikita Araslanov and Stefan Roth},
  title     = {Self-Supervised Augmentation Consistency for Adapting Semantic Segmentation},
  booktitle = {Proceedings of the IEEE/CVF Conference on Computer Vision and Pattern Recognition (CVPR)},
  pages     = {15384--15394},
  year      = {2021},
  publisher = {IEEE}
}

@inproceedings{hoyer2022daformer,
  author    = {Hoyer, Lukas and Dai, Dengxin and Van Gool, Luc},
  title     = {DAFormer: Improving Network Architectures and Training Strategies for Domain-Adaptive Semantic Segmentation},
  booktitle = {Proceedings of the IEEE/CVF Conference on Computer Vision and Pattern Recognition (CVPR)},
  pages     = {9924--9935},
  year      = {2022}
}

@article{liu2022deep,
  author    = {Liu, Xiaofeng and Yoo, Chaehwa and Xing, Fangxu and Oh, Hyejin and El Fakhri, Georges and Kang, Je-Won and Woo, Jonghye and others},
  title     = {Deep Unsupervised Domain Adaptation: A Review of Recent Advances and Perspectives},
  journal   = {APSIPA Transactions on Signal and Information Processing},
  volume    = {11},
  number    = {1},
  year      = {2022},
  publisher = {Now Publishers Inc.}
}

@inproceedings{ganin2015unsupervised,
  title={Unsupervised domain adaptation by backpropagation},
  author={Ganin, Yaroslav and Lempitsky, Victor},
  booktitle={International conference on machine learning},
  pages={1180--1189},
  year={2015},
  organization={PMLR}
}

@article{long2016unsupervised,
  author  = {Long, Mingsheng and Zhu, Han and Wang, Jianmin and Jordan, Michael I.},
  title   = {Unsupervised Domain Adaptation with Residual Transfer Networks},
  journal = {Advances in Neural Information Processing Systems (NeurIPS)},
  volume  = {29},
  year    = {2016}
}

@article{sener2016learning,
  title={Learning transferrable representations for unsupervised domain adaptation},
  author={Sener, Ozan and Song, Hyun Oh and Saxena, Ashutosh and Savarese, Silvio},
  journal={Advances in neural information processing systems},
  volume={29},
  year={2016}
}

@article{huang2021model,
  author  = {Huang, Jiaxing and Guan, Dayan and Xiao, Aoran and Lu, Shijian},
  title   = {Model Adaptation: Historical Contrastive Learning for Unsupervised Domain Adaptation without Source Data},
  journal = {Advances in Neural Information Processing Systems (NeurIPS)},
  volume  = {34},
  pages   = {3635--3649},
  year    = {2021}
}

@inproceedings{yeh2021sofa,
  author    = {Yeh, Hao-Wei and Yang, Baoyao and Yuen, Pong C. and Harada, Tatsuya},
  title     = {SOFA: Source-Data-Free Feature Alignment for Unsupervised Domain Adaptation},
  booktitle = {Proceedings of the IEEE/CVF Winter Conference on Applications of Computer Vision (WACV)},
  pages     = {474--483},
  year      = {2021}
}

@inproceedings{kundu2021generalize,
  author    = {Kundu, Jogendra Nath and Kulkarni, Akshay and Singh, Amit and Jampani, Varun and Babu, R. Venkatesh},
  title     = {Generalize Then Adapt: Source-Free Domain Adaptive Semantic Segmentation},
  booktitle = {Proceedings of the IEEE/CVF International Conference on Computer Vision (ICCV)},
  pages     = {7046--7056},
  year      = {2021}
}

@article{zheng2024360sfuda++,
  author    = {Zheng, Xu and Zhou, Peng Yuan and Vasilakos, Athanasios V. and Wang, Lin},
  title     = {360SFUDA++: Towards Source-Free UDA for Panoramic Segmentation by Learning Reliable Category Prototypes},
  journal   = {IEEE Transactions on Pattern Analysis and Machine Intelligence (T-PAMI)},
  year      = {2024},
  publisher = {IEEE}
}

@inproceedings{cordts2016cityscapes,
  author    = {Cordts, Marius and Omran, Mohamed and Ramos, Sebastian and Rehfeld, Timo and Enzweiler, Markus and Benenson, Rodrigo and Franke, Uwe and Roth, Stefan and Schiele, Bernt},
  title     = {The Cityscapes Dataset for Semantic Urban Scene Understanding},
  booktitle = {Proceedings of the IEEE Conference on Computer Vision and Pattern Recognition (CVPR)},
  pages     = {3213--3223},
  year      = {2016}
}

@article{zhang2024behind,
  title={Behind every domain there is a shift: Adapting distortion-aware vision transformers for panoramic semantic segmentation},
  author={Zhang, Jiaming and Yang, Kailun and Shi, Hao and Rei{\ss}, Simon and Peng, Kunyu and Ma, Chaoxiang and Fu, Haodong and Torr, Philip HS and Wang, Kaiwei and Stiefelhagen, Rainer},
  journal={IEEE Transactions on Pattern Analysis and Machine Intelligence},
  volume={46},
  number={12},
  pages={8549--8567},
  year={2024},
  publisher={IEEE}
}

@article{armeni2017joint,
  author  = {Armeni, Iro and Sax, Sasha and Zamir, Amir R. and Savarese, Silvio},
  title   = {Joint 2D-3D-Semantic Data for Indoor Scene Understanding},
  journal = {arXiv preprint arXiv:1702.01105},
  year    = {2017}
}

@inproceedings{yang2022source,
  author    = {Yang, Cheng-Yu and Kuo, Yuan-Jhe and Hsu, Chiou-Ting},
  title     = {Source-Free Domain Adaptation for Semantic Segmentation via Distribution Transfer and Adaptive Class-Balanced Self-Training},
  booktitle = {Proceedings of the IEEE International Conference on Multimedia and Expo (ICME)},
  pages     = {1--6},
  year      = {2022},
  publisher = {IEEE}
}

@inproceedings{zheng2023look,
  title={Look at the neighbor: Distortion-aware unsupervised domain adaptation for panoramic semantic segmentation},
  author={Zheng, Xu and Pan, Tianbo and Luo, Yunhao and Wang, Lin},
  booktitle={Proceedings of the IEEE/CVF International Conference on Computer Vision},
  pages={18687--18698},
  year={2023}
}

@inproceedings{yue2021prototypical,
  author    = {Yue, Xiangyu and Zheng, Zangwei and Zhang, Shanghang and Gao, Yang and Darrell, Trevor and Keutzer, Kurt and Vincentelli, Alberto Sangiovanni},
  title     = {Prototypical Cross-Domain Self-Supervised Learning for Few-Shot Unsupervised Domain Adaptation},
  booktitle = {Proceedings of the IEEE/CVF Conference on Computer Vision and Pattern Recognition (CVPR)},
  pages     = {13834--13844},
  year      = {2021}
}

@inproceedings{wang2020differential,
  title={Differential treatment for stuff and things: A simple unsupervised domain adaptation method for semantic segmentation},
  author={Wang, Zhonghao and Yu, Mo and Wei, Yunchao and Feris, Rogerio and Xiong, Jinjun and Hwu, Wen-mei and Huang, Thomas S and Shi, Honghui},
  booktitle={Proceedings of the IEEE/CVF conference on computer vision and pattern recognition},
  pages={12635--12644},
  year={2020}
}

@article{zhang2021transfer,
  author    = {Zhang, Jiaming and Ma, Chaoxiang and Yang, Kailun and Roitberg, Alina and Peng, Kunyu and Stiefelhagen, Rainer},
  title     = {Transfer Beyond the Field of View: Dense Panoramic Semantic Segmentation via Unsupervised Domain Adaptation},
  journal   = {IEEE Transactions on Intelligent Transportation Systems (T-ITS)},
  volume    = {23},
  number    = {7},
  pages     = {9478--9491},
  year      = {2021},
  publisher = {IEEE}
}

@inproceedings{yang2021capturing,
  author    = {Yang, Kailun and Zhang, Jiaming and Rei{\ss}, Simon and Hu, Xinxin and Stiefelhagen, Rainer},
  title     = {Capturing Omni-Range Context for Omnidirectional Segmentation},
  booktitle = {Proceedings of the IEEE/CVF Conference on Computer Vision and Pattern Recognition (CVPR)},
  pages     = {1376--1386},
  year      = {2021}
}

@inproceedings{zhao2024stable,
  author    = {Zhao, Dong and Wang, Shuang and Zang, Qi and Jiao, Licheng and Sebe, Nicu and Zhong, Zhun},
  title     = {Stable Neighbor Denoising for Source-Free Domain Adaptive Segmentation},
  booktitle = {Proceedings of the IEEE/CVF Conference on Computer Vision and Pattern Recognition (CVPR)},
  pages     = {23416--23427},
  year      = {2024}
}

@inproceedings{zou2019confidence,
  title={Confidence regularized self-training},
  author={Zou, Yang and Yu, Zhiding and Liu, Xiaofeng and Kumar, BVK and Wang, Jinsong},
  booktitle={Proceedings of the IEEE/CVF international conference on computer vision},
  pages={5982--5991},
  year={2019}
}

@inproceedings{luo2019taking,
  author    = {Luo, Yawei and Zheng, Liang and Guan, Tao and Yu, Junqing and Yang, Yi},
  title     = {Taking a Closer Look at Domain Shift: Category-Level Adversaries for Semantics Consistent Domain Adaptation},
  booktitle = {Proceedings of the IEEE/CVF Conference on Computer Vision and Pattern Recognition (CVPR)},
  pages     = {2507--2516},
  year      = {2019}
}

@inproceedings{wang2023balancing,
  title={Balancing logit variation for long-tailed semantic segmentation},
  author={Wang, Yuchao and Fei, Jingjing and Wang, Haochen and Li, Wei and Bao, Tianpeng and Wu, Liwei and Zhao, Rui and Shen, Yujun},
  booktitle={Proceedings of the IEEE/CVF conference on computer vision and pattern recognition},
  pages={19561--19573},
  year={2023}
}

@inproceedings{zheng2024semantics,
  author    = {Zheng, Xu and Zhou, Pengyuan and Vasilakos, Athanasios V. and Wang, Lin},
  title     = {Semantics Distortion and Style Matter: Towards Source-Free UDA for Panoramic Segmentation},
  booktitle = {Proceedings of the IEEE/CVF Conference on Computer Vision and Pattern Recognition (CVPR)},
  pages     = {27885--27895},
  year      = {2024}
}

@article{xie2021segformer,
  author    = {Xie, Enze and Wang, Wenhai and Yu, Zhiding and Anandkumar, Anima and Alvarez, Jose M. and Luo, Ping},
  title     = {SegFormer: Simple and Efficient Design for Semantic Segmentation with Transformers},
  journal   = {Advances in Neural Information Processing Systems (NeurIPS)},
  volume    = {34},
  pages     = {12077--12090},
  year      = {2021}
}

@article{cao2025cross,
  title={Cross-modal semantic transfer for point cloud semantic segmentation},
  author={Cao, Zhen and Mi, Xiaoxin and Qiu, Bo and Cao, Zhipeng and Long, Chen and Yan, Xinrui and Zheng, Chao and Dong, Zhen and Yang, Bisheng},
  journal={ISPRS Journal of Photogrammetry and Remote Sensing},
  volume={221},
  pages={265--279},
  year={2025},
  publisher={Elsevier}
}

@inproceedings{cao2023kt,
  title={Kt-net: knowledge transfer for unpaired 3d shape completion},
  author={Cao, Zhen and Zhang, Wenxiao and Wen, Xin and Dong, Zhen and Liu, Yu-Shen and Xiao, Xiongwu and Yang, Bisheng},
  booktitle={Proceedings of the AAAI conference on artificial intelligence},
  volume={37},
  number={1},
  pages={286--294},
  year={2023}
}
}


\end{document}